\documentclass[journal]{IEEEtran}
\usepackage{epsfig}
\usepackage{amsmath}
\usepackage{subfigure}
\usepackage{subfig}
\usepackage{epstopdf}
\usepackage{diagbox}
\usepackage{colortbl}
\usepackage{url}
\usepackage{multirow}
\usepackage{caption}
\usepackage{threeparttable}
\usepackage{pdfpages}
\usepackage{graphicx}
\usepackage{booktabs}
\usepackage{cite}
\usepackage{pmat}
\usepackage{mathrsfs}
\usepackage{amsmath}
\usepackage{array}
\usepackage{tabularx}
\ifCLASSINFOpdf
\else

\fi

\begin{document}

 \title{Deep Adaptive Proposal Network for Object Detection in Optical Remote Sensing Images}
\author{Lin~Cheng,
Xu~Liu,~\IEEEmembership{Student~Member,~IEEE,}
Lingling~Li,~\IEEEmembership{Member,~IEEE,}
Licheng~Jiao,~\IEEEmembership{Fellow,~IEEE,}
Xu~Tang,~\IEEEmembership{Member,~IEEE}
\thanks{This work was supported in part by the Major Research Plan of the National Natural Science Foundation of China (No. 91438201 and No. 91438103), the Fund for Foreign Scholars in University Research and Teaching Programs (the 111 Project) (No. B07048) \emph{(Corresponding author: Lingling Li)}}

\thanks{The authors are with the Key Laboratory of Intelligent Perception and Image Understanding of the Ministry of Education of China, International Research Center of Intelligent Perception and Computation, School of
Artificial Intelligence, Xidian University, Xi'an, China (e-mail: lchjiao@mail.xidian.edu.cn; chenglin\underline{\hspace{0.5em}}shine@163.com). }}
\maketitle

\begin{abstract}
Object detection is a fundamental and challenging problem in aerial and satellite image analysis. More recently, a two-stage detector Faster R-CNN is proposed and demonstrated to be a promising tool for object detection in optical remote sensing images, while the sparse and dense characteristic of objects in remote sensing images is complexity. It is unreasonable to treat all images with the same region proposal strategy, and this treatment limits the performance of two-stage detectors. In this paper, we propose a novel and effective approach, named deep adaptive proposal network (DAPNet), address this complexity characteristic of object by learning a new category prior network (CPN) on the basis of the existing Faster R-CNN architecture.
Moreover, the candidate regions produced by DAPNet model are different from the traditional region proposal network (RPN), DAPNet predicts the detail category of each candidate region. And these candidate regions combine the object number, which generated by the category prior network to achieve a suitable number of candidate boxes for each image. These candidate boxes can satisfy detection tasks in sparse and dense scenes.
The performance of the proposed framework has been evaluated on the challenging NWPU VHR-10 data set. Experimental results demonstrate the superiority of the proposed framework to the state-of-the-art.
\end{abstract}


\section{Introduction}
\IEEEPARstart{O}{bject} detection in optical remote sensing images is one of the hottest issues in many fields, such as disaster control, traffic monitoring, and traffic planning, has received increasing research interests in the fields of remote sensing image analysis over the last decades \cite{Cheng2016A, Han2018Advanced}.

The traditional object detection methods typically comprise three key procedures: candidate region proposal, feature extraction and classification. The latter can extract relatively less effective candidate boxes according to different rules \cite{uijlings2013selective, cheng2014bing, xu2014automatic}.
In \cite{uijlings2013selective}, selective search was proposed which combines the strength of both an exhaustive search and segmentation.
In \cite{cheng2014bing}, region proposal method called binarized normed gradient (BING) was proposed for object detection, which produce a small set of candidate object windows.
In \cite{xu2014automatic}, an iterative training method based on invariant generalized Hough transform was used to learn a robust shape model automatically. The model could capture the shape variability of the target contained in the training data set, and every point in the model is equipped with an individual weight according to its importance, which greatly reduces the false-positive rate.
Compared with the sliding window method, the above methods have greatly reduced the candidate region number and not affect the efficiency.

Conventional features adopted for object detection in remote sensing images are hand-crafted features such as color histograms, texture features, shape features histogram of oriented gradients (HOG) \cite{Dalal2005Histograms, Shao2012Car}, scale invariant feature transform (SIFT) \cite{Han2015Object, Lowe2004Lowe}, bag of words (BOW) \cite{Karklin2005A, Xu2010Object}, Saliency \cite{Zhang2014Saliency, Han2014Efficient}, and so on \cite{cheng2016survey}.

After feature extraction, a classifier can be trained using the training dataset by a number of possible approaches with the objective of minimizing the misclassification error. In practice, there are many different learning approaches including support vector machine (SVM) \cite{Xu2010Object}, AdaBoost \cite{Nguyen2008On, Leitloff2010Vehicle}, k-nearest-neighbor (kNN) \cite{Chen2011Sparse, Gong2013Automatic}, conditional random field (CRF) \cite{Zhong2007A}, sparse representation-based classification (SRC) \cite{Sun2011Automatic}, and artificial neural network (ANN) \cite{Park1991Electric}.

Currently, due to the powerful feature representation of deep learning, the convolutional neural networks (CNNs) has been used in classsification \cite{liu2018deep, jiao2017deep, zhao2017superpixel, jiao2016wishart, liu2016pol} and object detection. In particular, object detection using CNNs has achieved remarkable successes, such as two-stage detection approachs \cite{Girshick2013Rich, Girshick2015Fast, Ren2017Faster, Lin2016Feature} and single-stage detection methods \cite{Redmon2015You, Liu2015SSD}. For two-stage detection approachs, in order to locate and classify objects efficiently, R-CNN applied high-capacity convolutional neural networks (CNNs) to bottom-up region proposals that generated by selective search \cite{Girshick2013Rich}, which obtained good performance compared with traditional detection methods.
Fast R-CNN \cite{Girshick2015Fast} extends the previous work R-CNN \cite{Girshick2013Rich} by using the ROI pooling layer to deal with the problem of multi-stage pipeline, and thus improving the training and testing speed while also increasing detection accuracy. For further phrases, the RPN was proposed in Faster R-CNN \cite{Ren2017Faster} to replace the selective search, which generates multi-scale and translation-invariant region proposals based on the high-level features of images, for this reason Faster R-CNN achieved remarkable success on object detection.
And there are a sequence of advances \cite{Lin2016Feature} based on the two-stage framework.

For single-stage detection methods, YOLO \cite{Redmon2015You} unified the separate components of multiple bounding boxes regression and classification into a single neural network, greatly improved the speed of detection. And based on the unified detection architecture, the SSD \cite{Liu2015SSD} used multi-scale convolutional bounding box outputs attached to multiple feature maps at the top of the network to achieve better performance. To improve the detection accuracy of SSD, S$^3$FD \cite{Zhang2017S} developed the SSD by proposing a scale-equitable framework, to guarantee the enough number of match anchors for each object with different scales.
And there are a sequence of advances \cite{Fu2017DSSD, Shen2017DSOD} based on the single-stage SSD framework. Generally, the main advantage of the single-stage detectors is high efficiency, and the two-stage detectors are superior to the single-stage detectors in accuracy.


How to pursue the accuracy and the speed of detectors simultaneously, has always been a concern remain to be solved.
The RON \cite{Kong2017RON} answered this question by using the reverse connection to assists the former layers of CNNs with more semantic information, and guided the searching of objects with the objectness prior, which has obtained good performance reflected in speed and accuracy.
The RetinaNet \cite{Lin2017Focal} is another successful method that match the speed of previous single-stage detectors while surpassing the accuracy of two-stage detectors, it proposed the focal loss which applied a modulating term to the cross entropy loss in order to focus learning on hard examples and down-weight the numerous easy negative. The success of focal loss was largely attributed to overcome the extreme foreground-background class imbalance encountered during training of detectors, it proves that the imbalance between the foreground and background is a major problem that limits the performance of detectors, especially for dense detectors.

\begin{figure}[htb]
\centering
\begin{minipage}[b]{0\linewidth}
\centering
\centerline{\hspace{0cm}\epsfig{figure= 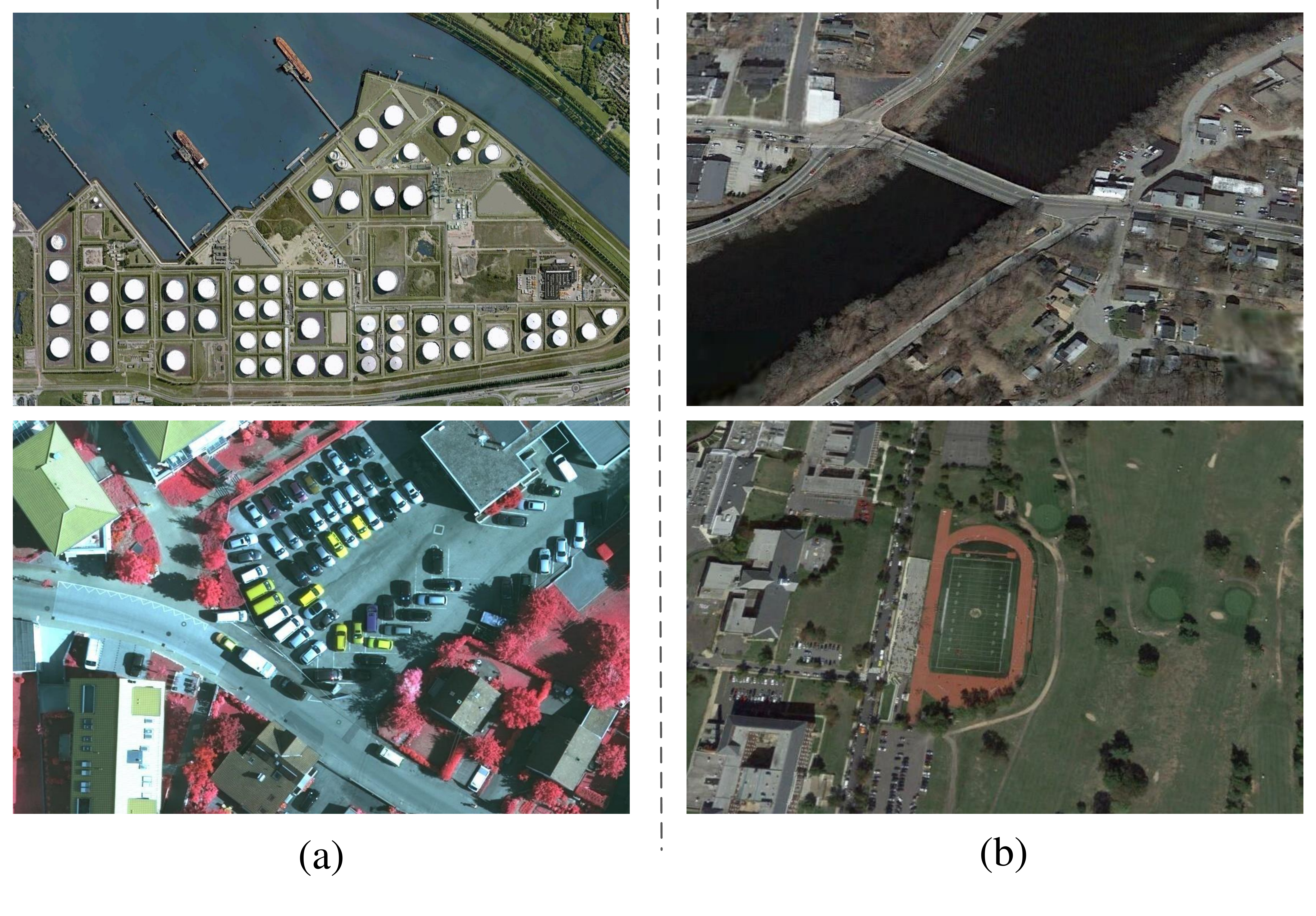,width=8.5cm}}
\end{minipage}
\caption{Some examples in the NWPH VHR-10 data set. (a) The typical examples of the dense objects (storage tank and vehicle). (b) The typical examples of sparse objects (bridge and ground track field).}
\label{s0}
\end{figure}

Driven by the success of focal loss \cite{Lin2017Focal} and S$^3$FD \cite{Zhang2017S}, guarantee plenty of positive and negative balance anchors for each object is a key point of improving the performance of detectors. To obtain enough anchors for each object, single-stage detectors generate anchors in each position of multiple layers of deep CNNs, and according to the overlaps between anchors and ground truth to select the positive and the negative anchors. However, without the region proposals, the detectors have to suppress too many negative anchors and regress the positive anchors, and it will increase the difficult to train the network with good performance.

Compared with single-stage detection methods \cite{Liu2015SSD, Lin2017Focal, Zhang2017S}, the two-stage detectors divided the object detection task into two sub-networks: the region proposal networks and the accuracy detection networks respectively. The region proposal networks rejected most of the negative samples, which ensure the balance of positive and negative anchors in a largely reduced searching space for accuracy detection networks. Besides, The twice regression generated accuracy final detection results. However, the two-stage detectors \cite{Ren2017Faster, Lin2016Feature, Cheng2016Learning, Li2017Rotation} fixed the candidate boxes to input the accuracy detection networks, while the overall arrangement of objects is different in different images, as illustrated in Fig. \ref{s0}, it is unreasonable to treat all image equality with fixed candidate boxes.

As mentioned above, detecting objects from high resolution aerial images is still a challenging task because the complex background information and uneven distribution of objects in remote sensing images, this challenge typically degenerated the performance of remote sensing images object detection.
In this paper, a novel and effective DAPNet is proposed to tackle above problem by learning a category prior network (CPN) and fine-region proposal network (F-RPN) to facilitate sparse and dense objects detection in remote sensing images.
The CPN is achieved by learning a new regression layer followed the convolution feature to automatically detect every class objects number relative to the given image.

For clarity, the main contributions of this paper are as follows:
\begin{itemize}
\item[1)]Focusing on the problem of sparse and dense objects detection in remote sensing images, such as ships and vehicles, and so on, we propose a effective DAPNet framework based on Faster-RCNN,
    which extensively reduce the redundancy boxes for accuracy detectors and detect multi-scale targets under complex background in remote sensing images, while significantly improve the performance with that of state-of-art approaches.

\item[2)]We develop a category prior network (CPN) to compute each category prior information via a small branch network, and model the process with a global ROI pooling layer followed by a binary classification layers and a regression layers. To adaptation the sparse and dense objects, we define nine levels base number as regression reference.

\item[3)]Aiming at the category balance characteristic of training data for accuracy object detection networks, we build a fine region proposal network (F-RPN) by change the classification branch and the testing strategies included in the existing RPN.  Combining the result of the CPN and F-RPN to achieve adaptive region proposals for each images, and thus facilitate sparse and dense objects detection in remote sensing images.
\end{itemize}

The remainder of this paper is organized as follows. Section \uppercase\expandafter{\romannumeral2} introduces the proposed DAPNet framework in detail. Experimental results and analysis are presented in Section \uppercase\expandafter{\romannumeral3}. Finally, Section \uppercase\expandafter{\romannumeral4} concludes this paper.

\begin{figure*}[htb]
\centering
\begin{minipage}[b]{0\linewidth}
\centering
\centerline{\hspace{0cm}\epsfig{figure= 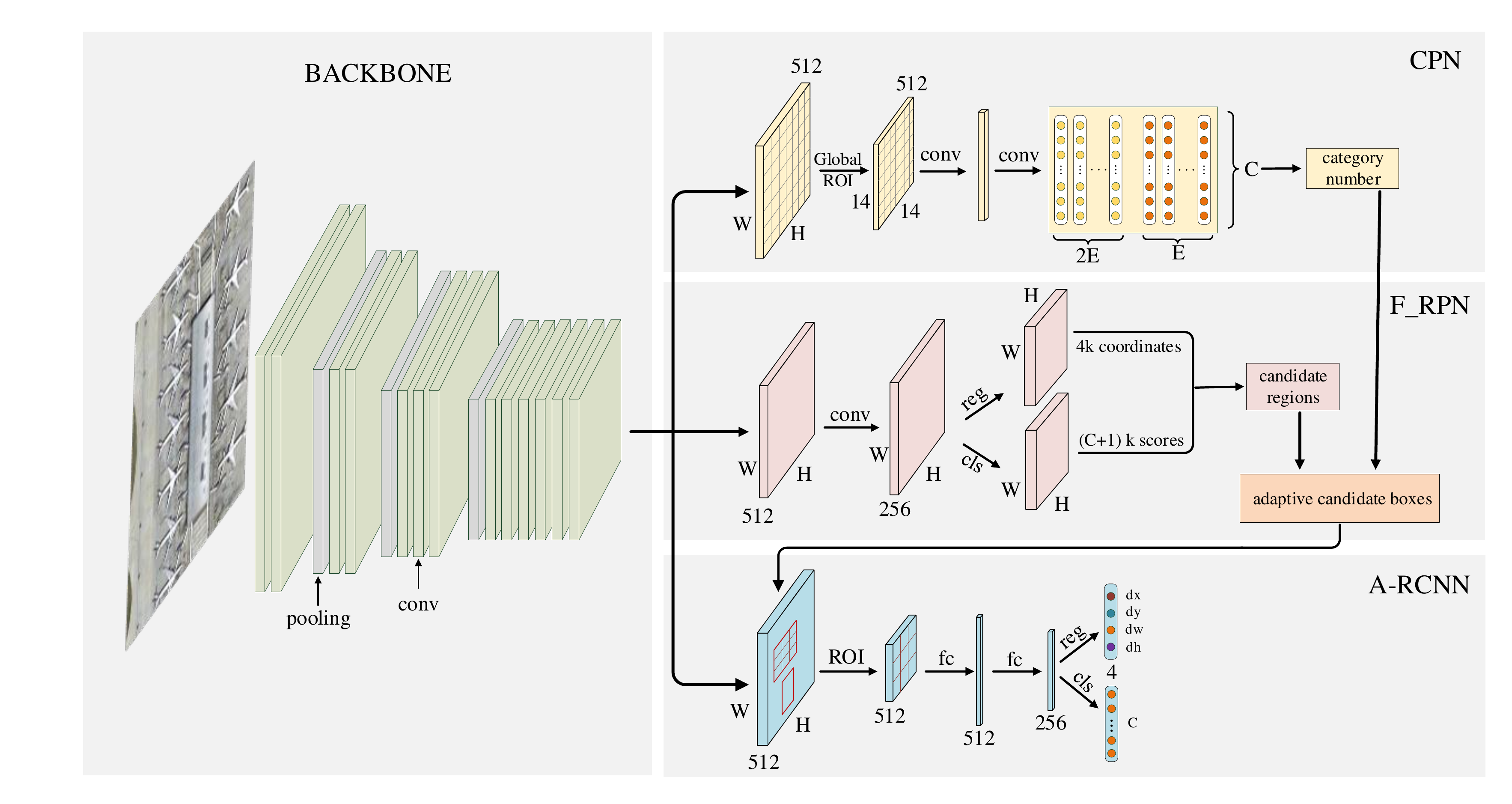,width=18.0cm}}
\vspace{0cm}
\end{minipage}
\caption{Overview of the proposed deep adaptive proposal network (DAPNet) framework. Given an input image, the backbone network was used to generate the high-level features representation of the image, then the CPN sub-network and the F-RPN sub-network followed the high-level features to obtain the category prior information and candidate regions for each image. Combination the results of the two sub-network to achieve adaptive region proposals, and finally the A-RCNN sub-network was used to classification and regression for each adaptive candidate boxes.}
\label{s1}
\end{figure*}

\section{Proposed Method}
\subsection{Overview of the Proposed Method}
The proposed object detection method, called DAPNet, is composed of four major components: the $VGG$-$16$ backbone network, the category prior network (CPN), the fine-region proposal network (F-RPN), and the accuracy detection network (A-RCNN), represented by aqua,
yellowish, pink, and wathet respectively, as represented in Fig. \ref{s1}. The DAPNet method is a novel network that can automatically adjust the number of candidate boxes according to the distribution of various objects in images. First, we use a deep fully convolutional backbone network to generate high-level convolutional features of the image. Then, the features are sent to three separate networks: the CPN, the F-RPN, and the A-RCNN,
in which the CPN predicts each category number for every image, and the F-RPN generates category independent possible candidate regions and classification results for each image. Finally, according to the each category number and candidate regions, we can produce adaptive candidate boxes, and perform an accurate object detection process to address these adaptive candidate boxes.

The overall architecture of $VGG$-$16$ network consists of sixteen weighted layers, including thirteen convolutional layers $conv1$, $conv2$, $conv3$, $conv4$, and $conv5$, two fully connected layers $fc6$ and $fc7$, and one softmax classification layer. Readers can refer to \cite{Simonyan2014Very} for more details. We build on the successful $VGG$-$16$ to learn our DAPNet model. In the training stage, the parameters of the first thirteen layers (thirteen convolutional layers) are transferred from $VGG$-$16$ weights which pretrained on $ImageNet$ $1000$-$class$ competition data set, and the last three layers are discarded.
Besides, to ensure that small objects have enough features for learning, we made some adjustments in the structure of VGG16 backbone network, we deleted the pool4 layer, and thus the high-level features $conv5$-$3$ has more information for small objects, as shown in the left of Fig. \ref{s1}. The light green layer represents the convolution layer, and the light grey layer indicates the pooling layer.
\subsection{Category Prior Network}
\begin{figure*}[htb]
\centering
\begin{minipage}[b]{0\linewidth}
\centering
\centerline{\hspace{0cm}\epsfig{figure= 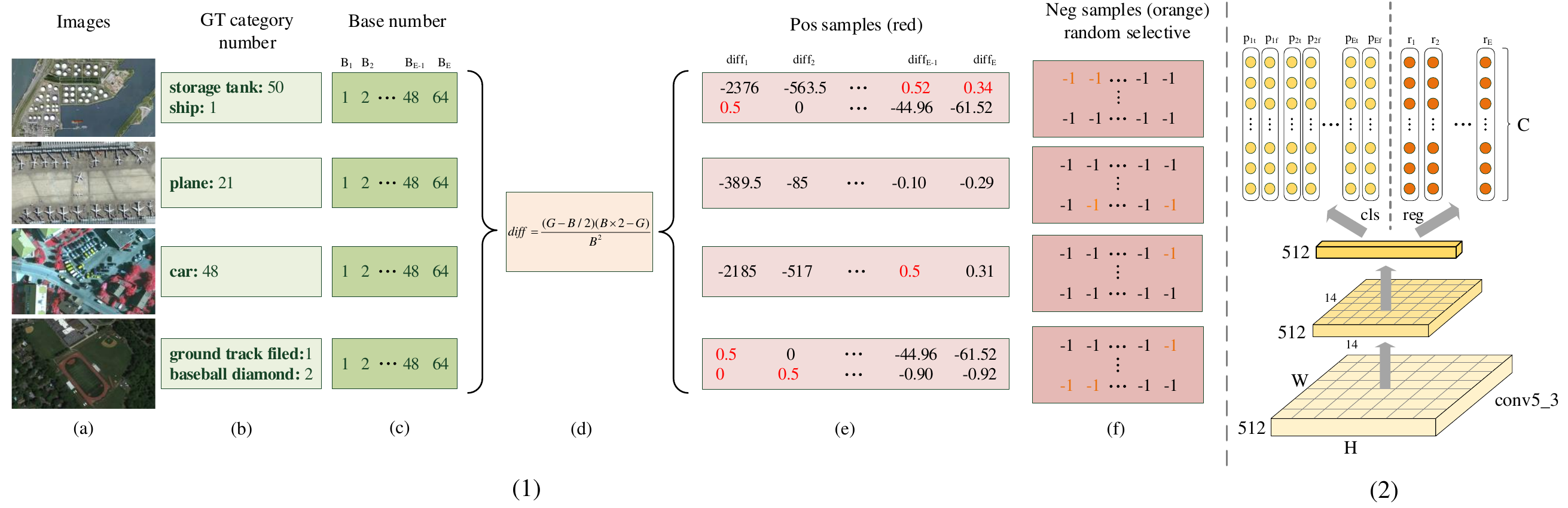,width=18.0cm}}
  \vspace{0cm}
\end{minipage}
\caption{(1) Multi-level number as the CPN regression reference. (a) Training images. (b) Ground-truth category numbers for each image. (c) $E$ levels base number for each image category number. (d) The formulation to calculate the difference between the ground truth category number with each level base number. (e) The value of difference results for each category level base number, and the positive regression reference (red numbers) according to the difference value. (f) Random selected negative regression reference according to the difference value. (2) The architecture of CPN.}
\label{s2}
\end{figure*}

A category prior network (CPN) takes an image (of any size) as input, and outputs each category prior information for the image as mentioned above. We are specifically inspired by the work of the Faster R-CNN \cite{Ren2017Faster}, which uses an RPN to generate multi-scale candidate regions. Our category prior network (CPN) generates multi-level numbers for each category similar to the structure of RPN. We model this process with a small fully convolutional network, which follows the backbone network. This small network includes a global ROI pooling layer, and a convolutional layer, followed by two sibling output layers, as shown in Fig. \ref{s2}(2).

To generate category number priors, we slide a small network over the high-level convolutional feature map output by the $conv5$-$3$ layer. First, the whole high-level convolutional feature map is mapped to an $n\times n$ feature map by a global ROI pooling layer. Then this small feature is fed into two convolutional layers, the first convolutional layer uses filters with $n\times n$ size, and the last convolutional layer produces $E$ regression values and 2 classification scores $p_{et}$ and $p_{ef}$ for each level category regression, with filters size $1\times 1$, and thus has a ($3\times E$)-channels output layer with $C$ object categories.

\subsubsection{Multi-Level Number as Regression Reference}
Object numbers can occur at any value because of the complex scene and uneven distribution of objects in remote sensing images. For each image high-level feature, we simultaneously predict $E$ levels regression number for each category, and the regression of each category does not affect each other.

Our design of CPN presents a novel scheme for addressing multi-level category number regression. In supervised batch learning, one of the keys to learn the category prior network is the definition of training samples. For each image, we predict $9$ levels (1, 2, 4, 8, 16, 24, 32, 48 and 64 numbers) as regression base references, as illustrated in Fig. \ref{s2}(1). First, the ground truth (GT) category numbers in each image are calculated and $9$ levels reference base number for each category are fixed. Then, the difference values for each category level are computed according to the following formula:
\begin{flalign}
diff_{ec} = \frac{(G_{c}^*-B_{e}/2)(B_{e}\times2-G_{c}^*)}{B_{e}^{2}}
\end{flalign}
in which $c$ represents the $c_{th}$ category, and $e$ means the $e_{th}$ regression level within the range of $E$. $G_{c}^*$ denotes the ground truth number for $c_{th}$ category, and $B_{e}$ is the $e_{th}$ level reference base number.
To facilitate different images and obtain the high-quality regression category numbers, the position of the difference value higher than 0 in every non-zero category number are recorded, and the category references at the position above are defined as the positive training samples. However, the ratio of $G_{c}^*$ to $B_{e}$ within $\frac1{4}$ to $\frac1{2}$ and 2 to 4 are ignored, in order to expand the disparity between the positive and negative training samples, and all of the rest category regression base numbers are defined as the negative searching space. The positive samples are shown the red color numbers in Fig. \ref{s2}(1)(e). Besides, to ensure the balance between the positive and negative numbers, we random select 3 times the number of positive samples in negative searching space as negative samples, as shown the orange color numbers in Fig. \ref{s2}(1)(f).

\subsubsection{Loss Function}
For training CPN, we assign a new regression and classification loss, we minimize an objective function following the multi-task loss in Fast R-CNN. Our multi-task loss function for an image is defined as
\begin{flalign}
L_{cpn}(\{p_{ec}\},\{r_{ec}\}) = \frac1{N_{cls}}\sum_{e=0}^E\sum_{c=0}^{C}L_{cls}(p_{ec},p_{ec}^*) &\nonumber\\
 +\alpha\frac1{N_{reg}}\sum_{e=0}^E\sum_{c=0}^{C+1}I_{ec}^*L_{reg}(r_{ec},r_{ec}^*)
\label{cnps:loss}
\end{flalign}
Here, $e$ is the level index of all reference level $E$, let $I_{ec}^*={0,1}$ be an indicator for matching the $c_{th}$ category $e_{th}$ base number to positive samples, it means if the $c_{th}$ category object exist in this image and the $e_{th}$ base number is responsible for this category. $p_{ec}$ is the predicted probability of $c_{th}$ category $e_{th}$ base number. The ground-truth label $p_{ec}^*$ is $1$ means that the image including $c_{th}$ category object and the $e_{th}$ base number is positive sample, and is $0$ represents the $c_{th}$ category $e_{th}$ base number is negative sample. The $r_{ec}$ is the regression result of $c_{th}$ category $e_{th}$ level, and $r_{ec}^*$ is that of ground truth regression value. The classification loss $L_{cls}$ is log loss over two class (object contain $p_{ect}$ or not $p_{ecf}$). $C$ represents the class number. For the regression loss, we use $L_{reg}(r_{ec},r_{ec}^*) = S(r_{ec}-r_{ec}^*)$ where $S$ is the robust function $SmoothL_{1}$ defined in \cite{Girshick2015Fast}. The term $I_{ec}^*L_{reg}$ means the regression loss is activated only for existing object category base number ($I_{ec}^*=1$) and is disabled otherwise ($I_{ec}^*=0$).

The classification and regression loss terms are normalized by $N_{cls}$ and $N_{reg}$ respectively, and weighted by balancing parameter $\alpha$. In our current implementation, $N_{cls}$ is the total classification numbers, including the positive and negative samples, about four times the number of objects. The $N_{reg}$ represents the total positive numbers for regression, about two times the number of objects. By default we set $\alpha=1$, which means that we bias toward better category number regression. For category number regression, similar to RPN, we regress the offsets of the category number.
\begin{flalign}
r_{ec} = log(G_{c}/B_{ec})  \\
r_{ec}^* = log(G_{c}^*/B_{ec})
\end{flalign}
where variables $G_{c}$, $G_{c}^*$, and $B_{ec}$ represents the predicted category number, ground truth category number, and base level category number respectively.
\subsubsection{Training and Testing CPN}
The CPN is a small network based on the output of backbone network, as mentioned before, the backbone network was pretrained on the $ImageNet$ $1000$-$class$ competition dataset \cite{Russakovsky2014ImageNet}. For the other convolutional layers in CPN, we initialize the parameters with the $xavier$ method \cite{Glorot2010Understanding}.
The CPN can be trained end-to-end by back-propagation and stochastic gradient descent (SGD) \cite{Lecun1989Backpropagation}, which includes iterative forward passes that takes labeled data as input. According to the above multi-task loss function, it is possible to optimize for each category number.

After trained the CPN, the training images were sent to the network and the offset of each image category level number was produced. However, the purpose of the CPN was to obtain the every category number, instead of all category levels number.
Therefore, during the testing process of CPN, we predicted scores for each category level regression. According to the output scores of CPN, we fixed a 0.7 threshold to filter the negative category and low score category levels regression, we only keep category levels regression which the scores higher than 0.7 threshold, and compute the average regression result as this category final predicted number result.

\subsection{Fine Region Proposal Network}
The advances in Faster R-CNN are driven by the success of region proposal network that extracting high-quality candidate regions and nearly cost-free. Typically, the existing RPN takes an image as input and produces a fixed collection of rectangular object boxes with different scales and aspect ratios. In order to produce the candidate regions, the binary classification scores were used to filter the negative rectangular object boxes, and the top fixed ranked rectangular object boxes were selected as candidate regions. Yet the number of objects in different images is not necessarily the same, it is unreasonable to select the same fixed candidate regions for all images. Therefore, we proposed a fine region proposal network (F-RPN) to facilitate the adaptive region proposals for different images.

\begin{figure}[htb]
\centering
\begin{minipage}[b]{0\linewidth}
\centering
\centerline{\hspace{0cm}\epsfig{figure= 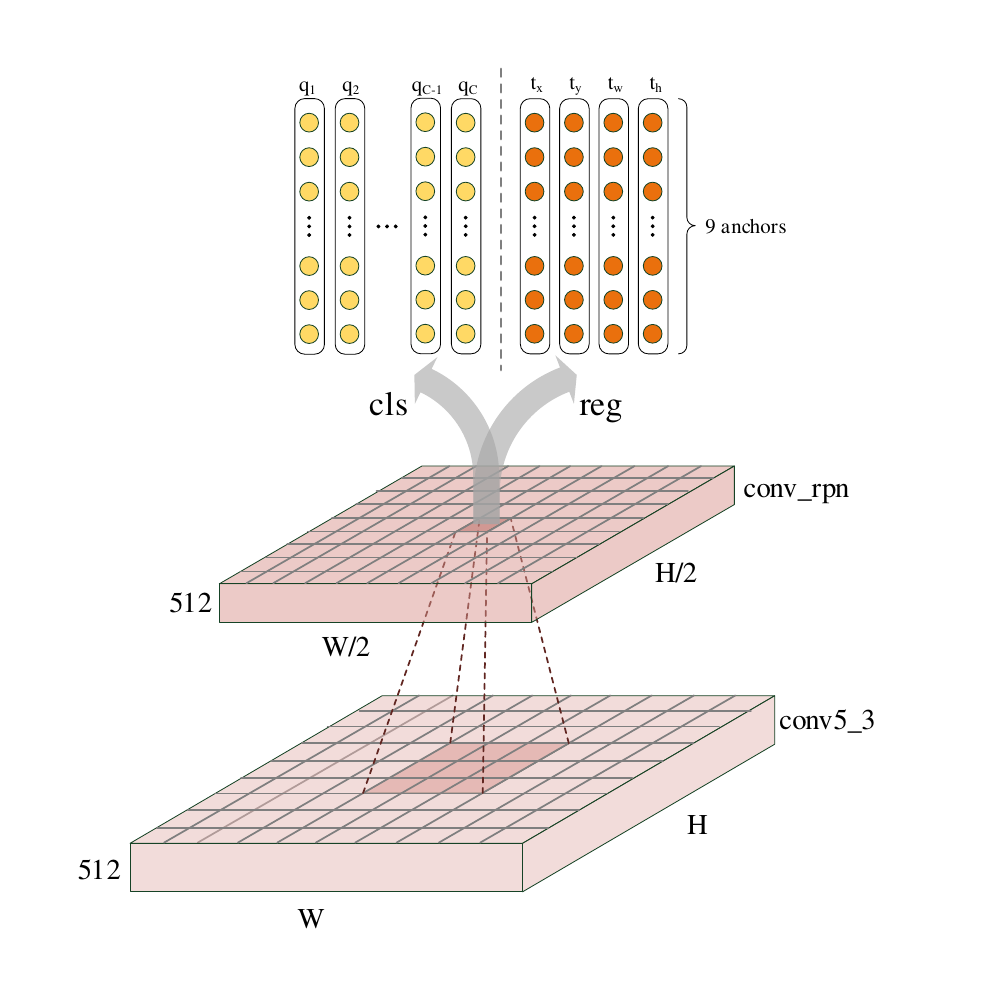}}
\end{minipage}
\caption{Illustration of F-RPN architecture, which generates fine class for each candidate regions.}
\label{s3}
\end{figure}

Similar to RPN, the architecture of F-RPN is based on a $VGG$-$16$ backbone network to extract high-level semantic features from the image. The F-RPN consists of a $n\times n$ convolutional layer $conv$-$rpn$ and two sibling $1\times 1$ convolutional output layers, the one for boxes fine classification and other for boxes regression. In particular, we designed a novel proposal selective strategy and a fine classification network for generating more effective and comprehensive high-quality region proposals. The overall architecture of F-RPN was shown in the center right of Fig. \ref{s3}.

Before training the fine classification and regression network, the positive and negative anchor training samples was required to determine. For every position in $conv$-$rpn$ feature maps, we generated $12$ anchors of four scales ($96\times 96$, $128\times 128$, $256\times 256$, and $512\times 512$ pixels) and three aspect ratios ($1:1$, $2:1$, and $1:2$). We assigned the anchors which $IoU$ value higher than 0.5 with positive label that corresponding to its category. For negative anchor samples, to guarantee the balance between the positive and negative anchor training samples, we randomly selected three times number of positive anchors from the $IoU$ overlap with all ground truth box lower than $0.3$, and assigned the negative label with label $0$ to these anchors.

\subsubsection{Fine Classification Network}
The major challenge for object detection is accurate object localization, and the one problem is the imbalance region proposals for each object which limits the performance of object localization. An important property of our approach is that it has the CPN and the F-RPN, and thus we can balance the region proposal numbers for every image. Therefore, we designed the fine classification network that derived from the RPN but is extended to handle multi object categories, which determines the detail class of an anchor and its confidence. The training objective for fine classification network is the softmax loss over all positive and negative anchor samples:
\begin{flalign}
L_{cls}(q,I) = \frac1{M}(\sum_{i\in Pos}^{M}-I_{i}^{*}log(\hat{q}_{i}^{c^{*}}) + \sum_{i\in Neg}-log(\hat{q}_{i}^{0}))
\label{cnps:rpnclsloss}
\end{flalign}
where
\begin{flalign}
\hat{q}_{i}^{c} = \frac{exp(q_{i}^{c})}{\sum_{c} exp(q_{i}^{c})}
\label{cnps:rpnclsloss}
\end{flalign}
in which $q_{i}$ is the predicted softmax outpus and the $q_{i}^{c}$ denotes the corresponding predicted $c_{th}$ category scores, $q_{i}^{c^{*}}$ represents the predicted scores for $i_{th}$ samples ground truth $c_{th}^{*}$ category, the class 0 represents the background. $I_{i}^{*}$ is an indicator for matching the $i_{th}$ candidate regions to the ground truth boxes, if matched $I_{i}^{*}=1$ and otherwise $I_{i}^{*}=0$. $M$ is the number of matched candidate regions. The fine classification network is nearly cost-free compared with the original classification network in RPN, after obtained the class for each region proposal, we used a category NMS, the method will be explain in the testing strategy section in detail.

\subsubsection{Regression Network}
The architecture of regression network is inherited from the RPN in Faster-RCNN. To regress the offset between the positive anchor boxes and the corresponding ground truth boxes, we slide a convolutional layer over the $conv$-$rpn$ to implement the regression purposes. Each anchor is represented by four parameters of center point $x$-$coordinate$, center point $y$-$coordinate$, anchor width, and anchor height. The robust function $SmoothL_{1}$ is used to calculate the difference between the output of regression network and the offset of anchors with corresponding ground truth boxes. Therefore, the training loss for single image is the average value of all positive anchors:
\begin{flalign}
L_{reg}(t^{u},t^{v}) = \frac1{M}\sum_{i\in(x,y,w,h)}SmoothL_{1}(t_{i}^{u}-t_{i}^{v})
\label{cnps:rpnregloss}
\end{flalign}
in which $u$ represents the index of the positive sample, $t_{i}^{v}$ is the corresponding offset between the $i_{th}$ anchor with the ground truth boxes, $t_{i}^{u}$ is the predicted offset. And where $x$, $y$, $w$, $h$ are the center $x$-$coordinates$, center $y$-$coordinates$, width of boxes, and height of boxes, respectively. Where the $SmoothL_{1}$ is a robust $L_{1}$ loss that is less sensitive to outliers compared with the $L_{2}$ loss. It is uncomplicated for us to train the F-RPN network, while the challenge is how to select the high-quality region proposals from all predicted boxes, we will illustrate the testing strategy in section.

\subsubsection{Training Objective}
The F-RPN can be trained end-to-end by a multi-task loss function that including the regression loss and the fine classification loss as mentioned above. we defined a weight term $\beta$ to balance the regression network and classification network in F-RPN:
\begin{flalign}
L_{frpn}(q,I,t^{u},t^{v}) = L_{cls}(q,I) + \frac1{\beta} L_{reg}(t^{u},t^{v})
\label{cnps:rpnallloss}
\end{flalign}
where $L_{frpn}(q,I,t^{u},t^{v})$ is the loss to train the F-RPN. In experiment, we set the $\beta=0.5$, which means that we bias toward better box location.

\subsubsection{Testing Strategy}
After trained the F-RPN network, all training images were sent to the network to product the enough amount of region proposals for each image, the next problem is to select high-quality proposals that contain the object. Faster R-CNN \cite{Ren2017Faster} deals with the problem by threshold filter and NMS to select $2000$ region proposals for each image, while the NMS that according to binary scores is unreasonable, because of the predicted scores for different class of object is inequable. Beside, the amount of region proposals for each image is too much,
actually there are not so many objects in each image that need to be detected. Therefore, we extended the testing strategy based on the RPN, the details are introduced as follows.

\textbf{Scores Filter:} The output of F-RPN includes two items, the one is the region proposals, and the other is the category scores for each region proposal. we filter the region proposals that do not contain any object according to the scores, if the predict scores for a region proposal is the class $0$, it means this region proposal is background, then the equivalent of the region proposal was filtered. Beside, we filter all predicted region proposals that do not contained in the image.

\textbf{Category NMS:} Generally, the object information of single image is different, and the category scores for every box was obtained by the fine classification network. Therefore, we propose the category NMS strategy to filter redundant region proposals and keep the high-quality region proposals.
It is easy to realize the category NMS, according to the scores for every category, we set different value of threshold that basis on the category number to filter single category redundant boxes. To a certain degree, the category NMS can reduce the interaction between classes.

At test time, after the CPN prediction, the category numbers for each image is produced, this prior information is the key point to select the candidate regions and reduce the computation. Due to the category number in a single image is far less than $2000$ obviously, most of the region proposals is redundant.
And thus, the number of each category retained in a image is in accordance with the category number and the output results of category NMS.
In short, the fine-region proposal networks can generate the adaptive and high-quality candidate boxes.

\subsection{Accurate Object Localization}
The existence of CPN and F-RPN is to generate high-quality adaptive candidate boxes, and the candidate boxes are the area that most likely contains the object. Therefore, in order to achieve the target of object detection, the A-RCNN is used to realize the regression and classification of this candidate boxes.
To implement the A-RCNN, the candidate boxes were mapped to the high level feature $conv5$-$3$, and then we used the ROI pooling to divide each proposal feature into a fixed $s\times s$ size, followed by two fully connected layers to classify and regress this object, in our experiments, we set $s=3$, as shown in the lower right of Fig. \ref{s1}. The loss of AR-CNN subnet is the same as that of Fast R-CNN \cite{Girshick2015Fast}, we do not elaborate too much here.

The purpose of the CPN, the F-RPN and the A-RCNN are to strengthen the feature of the object and restrain the background characteristics, and thus the three subnets are based on the output of backbone high level feature $conv5$-$3$, it allows for sharing convolutional layers between the three network, rather than learning three separate networks.
In the training phase, the loss of DAPNet for each image is the combination of three subnets loss as mentioned earlier, and the loss weight of each subnet was set to 1.
During the test process, after the accurate object proposals were obtained, we used category NMS to get the final object localization, combined with the result of classification scores, the final detection results were produced.

\section{Experiments}
For the experiments in this paper, NWPU VHR-10 geospatial object detection dataset is used to investigate the performance of our proposed DAPNet framework. We first introduce the data set and the evaluation metrics for the experiments. Then, the implementation details of the DAPNet is presented, which including the analysis of hyper-parameters and a ablation experiment. Finally, the proposed DAPNet method is compared with several traditional methods and deep learning methods, including the results presentation and numerical analysis.

\subsection{Description of Data Set}
The NWPU VHR-10 \cite{Cheng2014Multi} is one of the pioneering works in remote sensing object detection filed, which is designed to provide a standardized test bed for multi-class object detection in remote sensing images. This data set was cropped from Google Earth and Vaihingen dataset and then manually annotated by experts, it contains ten classes of objects, are airplane, ship, storage tank, baseball diamond, tennis court, basketball court, ground track field, harbor, bridge, and vehicle respectively. In particular, the data set contains sparse and dense objects, in which the property of sparse and dense is according to the object distributions in images. For instance, the part of storage tanks and vehicles are dense objects, while the ground track fields are sparse object. The ten classes object statistics of the NWPU VHR-10 data set is shown in Table \ref{tab:nwpu}.

\renewcommand\arraystretch{1}
\begin{table}[htbp]
\centering
\captionsetup{justification=centering}
\caption{\\THE ANALYSIS OF THE TRAINING DATASET}
\label{tab:nwpu}
\centering
\begin{tabularx}{7.5cm}{cXXX}
    \hline
    \hline
    Class Name & Total Number & Minimum Number & Maximum Number\\
    \hline
    Airplane & 757 & 1 & 31 \\
    Ship & 302 & 1 & 15 \\
    Storage tank & 655 & 6 & 63 \\
    Baseball diamond & 390 & 1 & 8 \\
    Tennis court & 524 & 1 & 24 \\
    Basketball court & 159 & 1 & 6 \\
    Ground track field & 163 & 1 & 1 \\
    Harbor & 224 & 3 & 18 \\
    Bridge & 124 & 1 & 5 \\
    Vehicle & 477 & 2 & 48 \\
    \hline
    \hline
\end{tabularx}
\end{table}
\normalsize

Table \ref{tab:nwpu} shows the object distribution of the NWPU VHR-10 dataset, including the total object number, the minimum number, and the maximum number for each category in each image. It is obvious that the data set contains the sparse and dense object in different images, and thus its suitable for the evaluation of our proposed method. The part image of the NWPU VHR-10 data set is shown in Fig. \ref{s0}.

The NWPU VHR-10 data set includes two parts: 650 positive images and 150 negative images, totally 800 images for training our DAPNet framework and comparative trials. First, we randomly select 500 images from the collection of positive images, combining with 150 negative images to form the training dataset.
And the rest 150 positive images is used to test the performance of the object detection methods in this paper.

\subsection{Evaluation Metrics}
The mean Average Precision (mAP) and the Precision-recall curves (PRC) are adopted to compare the detection performance of different approaches. To allow a better understanding of the mAP, we first explain the PRC and the Average precision (AP).

\subsubsection{Precision-Recall Curves}
The output of the object detection framework is a collection of bounding boxes, and the bounding boxes that the overlap with ground truth higher than 0.5 are supposed to the positive samples, while the others are considered as the negative samples. In addition, if several detections overlap with a same ground-truth bounding box, only one is considered as true positive, and the others are supposed as false negatives.

The precision indicator measures the proportion of detections that are true positives, and the recall indicator represents the fraction of positives that are correctly identified. The precision and recall indicators are formulated as follows:
\begin{flalign}
Precision = \frac{TP}{(TP+FP)}
\label{evaluation:Precision}
\end{flalign}
\begin{flalign}
Recall = \frac{TP}{(TP+FN)}
\label{evaluation:recall}
\end{flalign}
where, TP, FP, and FN denote the number of true-positive, the number of false-positive, and the number of false-negative respectively. The precision-recall curves takes the recall as abscissa and the precision as ordinate. If the detection method can keep high value of precision with the increasing of recall, it means the good performance of the detection approach.

\subsubsection{Average Precision}
The average precision (AP) computes the average value of the precision over the interval from $recall = 0$ to $recall = 1$, i.e, it can be formulated as:
\begin{flalign}
AP = \int_0^1 p(r)dr
\label{evaluation:meanap}
\end{flalign}
where, the $p$ denotes the value of precision and the $r$ is the value of recall. Hence, the AP is equal to taking the area under the curve, the higher the AP, the better the performance, and the mean AP represents the average value of all category AP.

\subsection{Implementation Details}

\begin{figure}[htb]
\centering
\begin{minipage}[b]{0\linewidth}
\centering
\centerline{\hspace{0cm}\epsfig{figure= 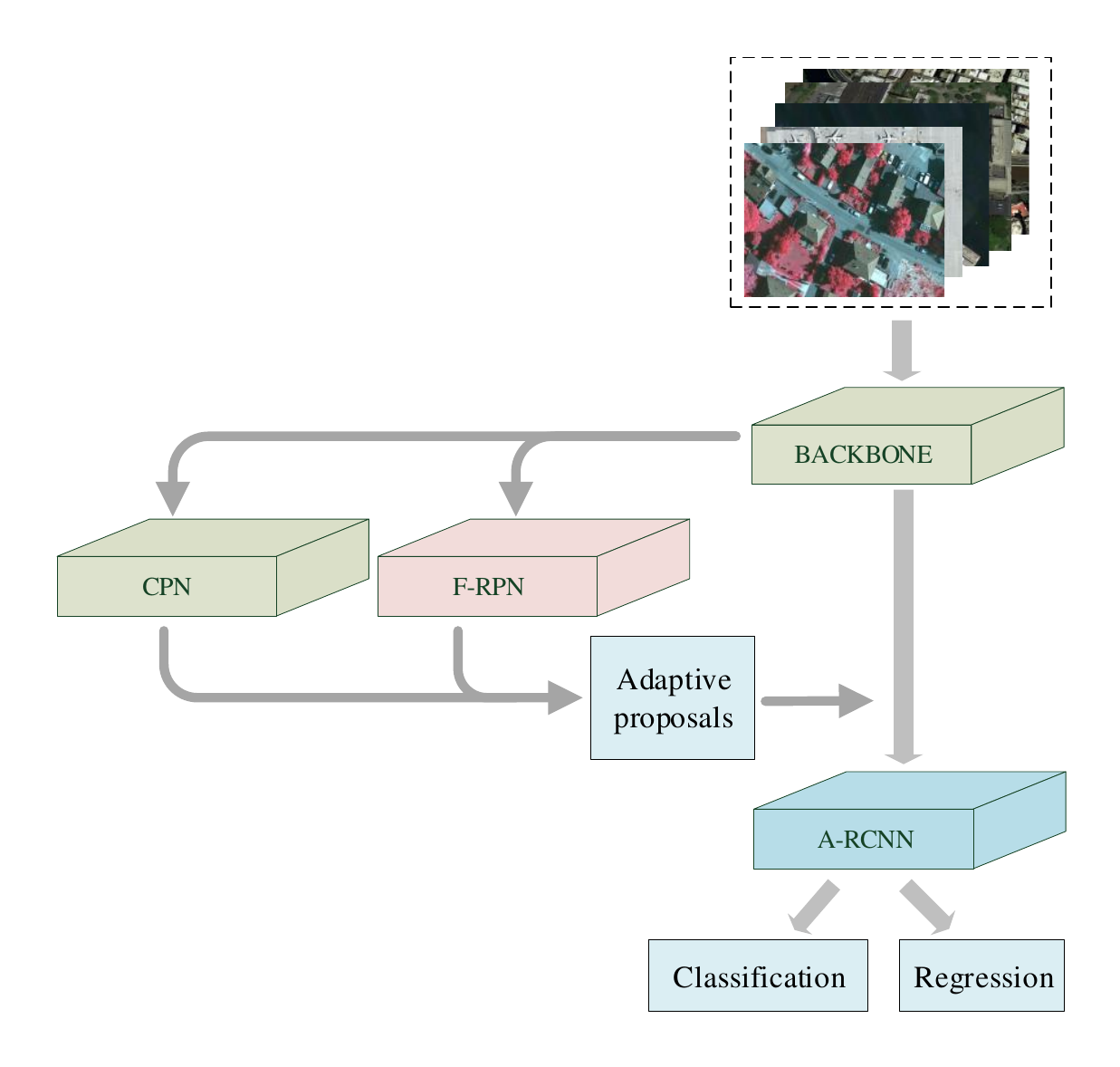, width=8cm}}
\end{minipage}
\caption{DAPNet is a single, unified network for object detection. The F-RPN and CPN serves as the prior information of this unified network.}
\label{s4}
\end{figure}

We run a considerable amount of experiments to analyze the behavior of the DAPNet framework with various hyper-parameters, including the contribution of the CPN network with different level references, and the commitment of F-RPN network with various anchor scales. For fair comparisons with other methods, we use the same images and data enhancement strategy for all network training and testing.

\subsubsection{Baselines and Network Initialization}

\begin{figure*}[htb]
\centering
\begin{minipage}[b]{0\linewidth}
\centering
\centerline{\hspace{0cm}\epsfig{figure= 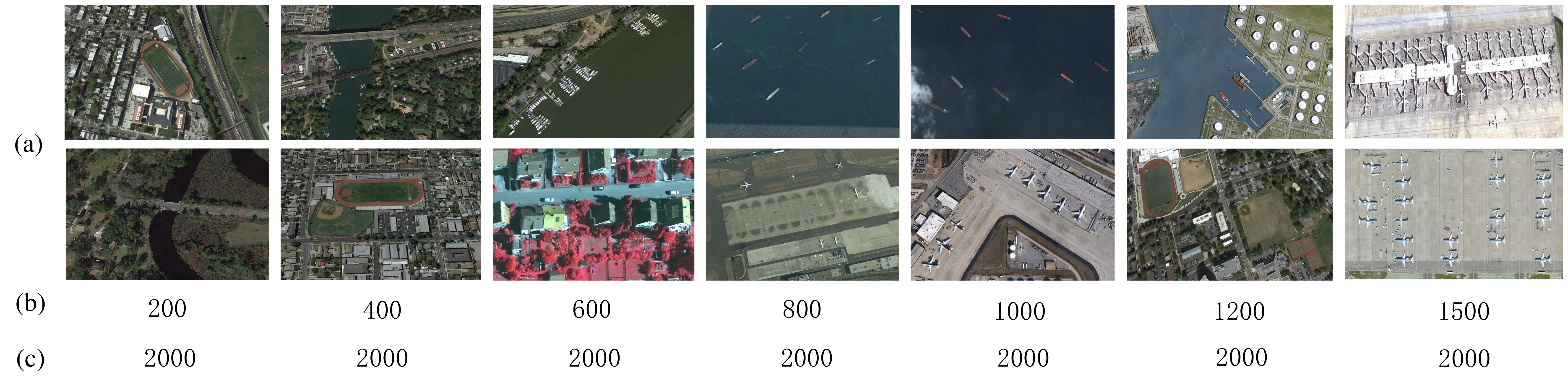,width=18.0cm}}
  \vspace{0cm}
\end{minipage}
\caption{Comparison of the region proposal number for the F-Faster R-CNN and DAPNet methods. (a) Images. (b) Adaptive proposal numbers for the DAPNet method. (c) Proposal numbers for the F-Faster R-CNN method.}
\label{s9}
\end{figure*}

For comprehensive reveal the superiority of our proposed DAPNet methods, we carry out comparative experiments with several baselines. Our DAPNet is inspried by Faster R-CNN \cite{Ren2017Faster}, so we directly use Faster R-CNN as the baselines, marked as Faster R-CNN.
Then, we exhibit the influence of fine-region proposal network and the CPN network, marked as F-Faster R-CNN. Besides, there are several baselines to compare with our DAPNet framework, including the traditional detectors FDDL, COPD and the deep leraning methods transferred CNN, RICNN \cite{Cheng2016Learning}.

For network initialization, there are two ways to initialize network weights in the training process: one is a random initialization using NWPU VHR-10 data set and the other is a fine-tuned initialization by using the trained CNN models on a publicly data set. As is common practice, all network backbones are pre-trained on the $ImageNet$ $1000$-$class$ classification dataset \cite{Russakovsky2014ImageNet} and then fine-tuned on our detection dataset. The training and testing process of our DAPNet are performed using the open source Caffe \cite{Jia2014Caffe} framework.

\subsubsection{Training and Testing Strategy}
The experiment consists of two procedures: the training process and the testing process, both of them are conducted on a computer with an NVIDIA Titan X GPU, 64 GB of memory, and the Compute Unified Device Architecture (CUDA) to improve the speed.

For the detection network in remote sensing images, we describe a novel DAPNet method that learns a unified network composed of CPN, F-RPN and A-RCNN with shared convolutional layers. We use the approximate joint training strategy to train our proposed method DAPNet, the CPN, F-RPN and A-RCNN are merged into one network during training as in shown Fig. \ref{s4}. In each stochasitc gradient descent (SGD) iteration, during the process of forward propagation, CPN generates category priors, which combine with the output of F-RPN to produce the adaptive region proposals. Then, the adaptive region proposals are used to train the A-RCNN detector. The backward propagation takes place as usual, where for the shared layers the backward propagated signals from all the CPN loss, the F-RPN loss and the A-RCNN loss are combined. This solution is easy to implement, but it ignores the derivative w.r.t the proposal boxes coordinate that are also network responses, so is approximate. In our experiments, in order to ensure fairness, the Faster R-CNN and F-RPN are both trained in this way. The hyper-parameters of these networks are defined in Table \ref{tab:arguments}.

\begin{table}[htbp]
\captionsetup{justification=centering}
\caption{\\ARGUMENTS FOR TRAINING DAPNet}
\label{tab:arguments}
\centering
\begin{tabular}{ll}
    \hline
    \hline
    Argument & Value\\
    \hline
    Learning rate & 0.001(0.0001 for finetune) \\
    Batch size & 1(for F-RPN and CPN) \\
    Dropout & 0.5 \\
    Momentum & 0.9 \\
    Weight decay & 0.0005 \\
    Max iter & 50000 \\
    \hline
    \hline
\end{tabular}
\end{table}
\normalsize

\subsubsection{Category Prior with CPN}
In the training processing, to achieve better performance in our category prior network. In the experiments, 9 levels base number (1, 2, 4, 8, 16, 24, 32, 48, 64) are set as regression references, it means that there are enough regression range for each category, to facilitate the dense and sparse objects in remote sensing images. And we assign the category levels which difference value higher than 0 as positive level number, the ratio of groundtruth number to base number between the $\frac1{4}$ to $\frac1{2}$ and 2 to 4 are assigned as ignored level numbers to avoid error regression. The rest of all category level are assigned as negative level numbers. During the test process, we only trust the regression results that whose category level scores is higher than 0.5. For each category, we calculate the average result of this convincing regression results, as the final prediction prior result.

To investigate the behavior of our CNPNs, we conduct a ablation study. To verify the contribution of CPN on the final detection precision, the proposed network DAPNet is compared with the F-RPN without the CPN, the results are represented in Table \ref{tab:threesome}. It is obvious that the DAPNet method has better performance on the NWPU VHR-10 data set, especially the category of storage tank and vehicle.

\begin{figure}[htb]
\centering
\begin{minipage}[b]{0\linewidth}
\centering
\centerline{\hspace{0cm}\epsfig{figure= 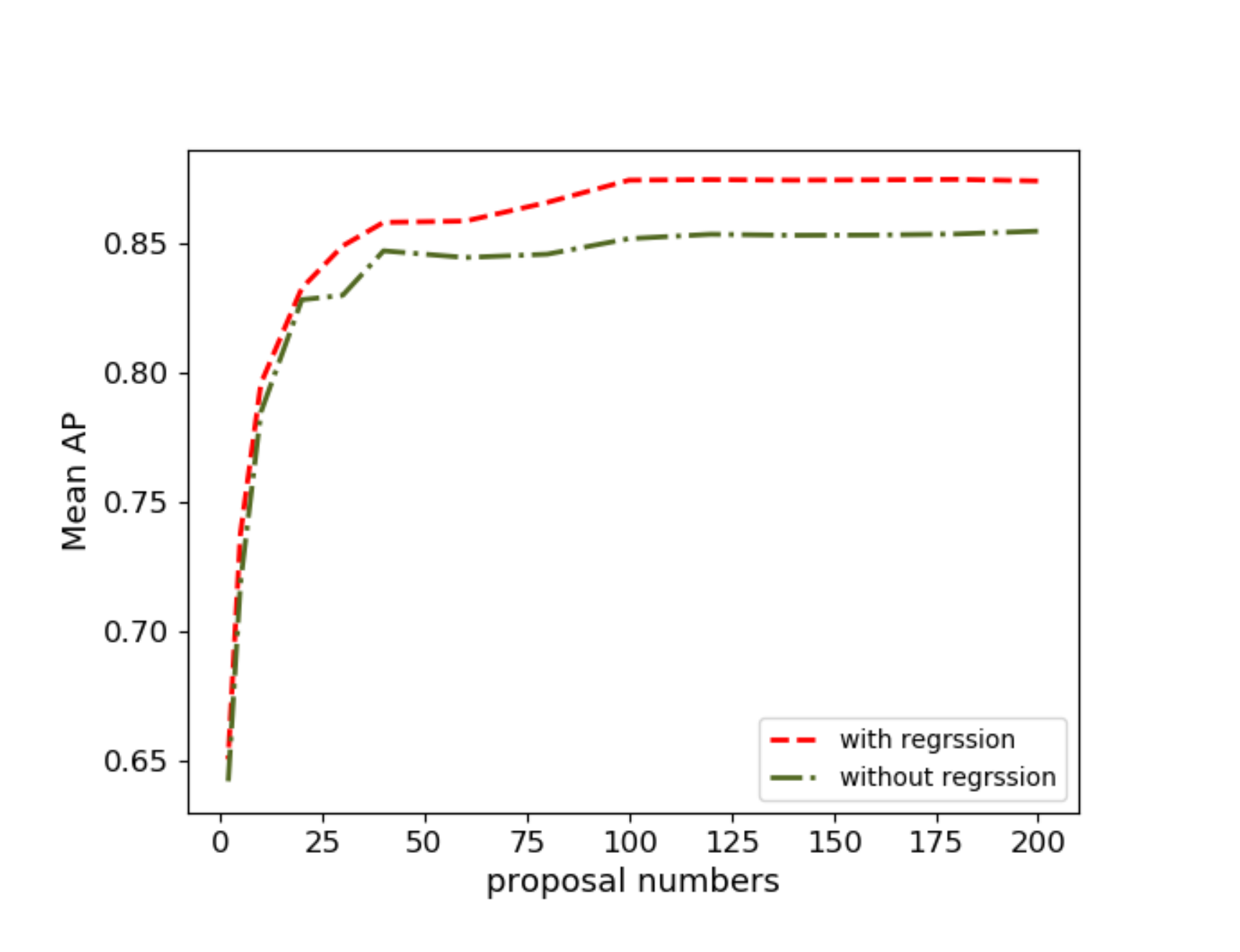, width=7.5cm}}
\end{minipage}
\caption{Mean AP versus numbers of proposals on the NWPU VHR-10 test data set (with $IoU$=0.5).}
\label{s8}
\end{figure}

Besides, to perform a further experiment, we disentangle the influence of CPN on detection speed. As we know, the key point that limits the speed of two-stage detection framework is the number of region proposals, and the dense object requires more region proposals compared with the demands of the spare object. For more intuitive, the output region proposal numbers of F-Faster R-CNN and DAPNet are extracted and observed, as shown in Fig. \ref{s9}.

Fig. \ref{s9} shows the region proposal number of our proposed F-Faster R-CNN and DAPNet, and Table \ref{tab:threesome} shows the detection results of proposed DAPNet method and F-Faster R-CNN. it can be seen that the proposed DAPNet algorithm demonstrates better overall detection performance compared with F-Faster R-CNN, while demands less region proposals. In addition, the DAPNet framework generates adaptive region proposals according to the object distribution in images.

\begin{figure*}[htb]
\centering
\begin{minipage}[b]{0\linewidth}
\centering
\centerline{\hspace{0cm}\epsfig{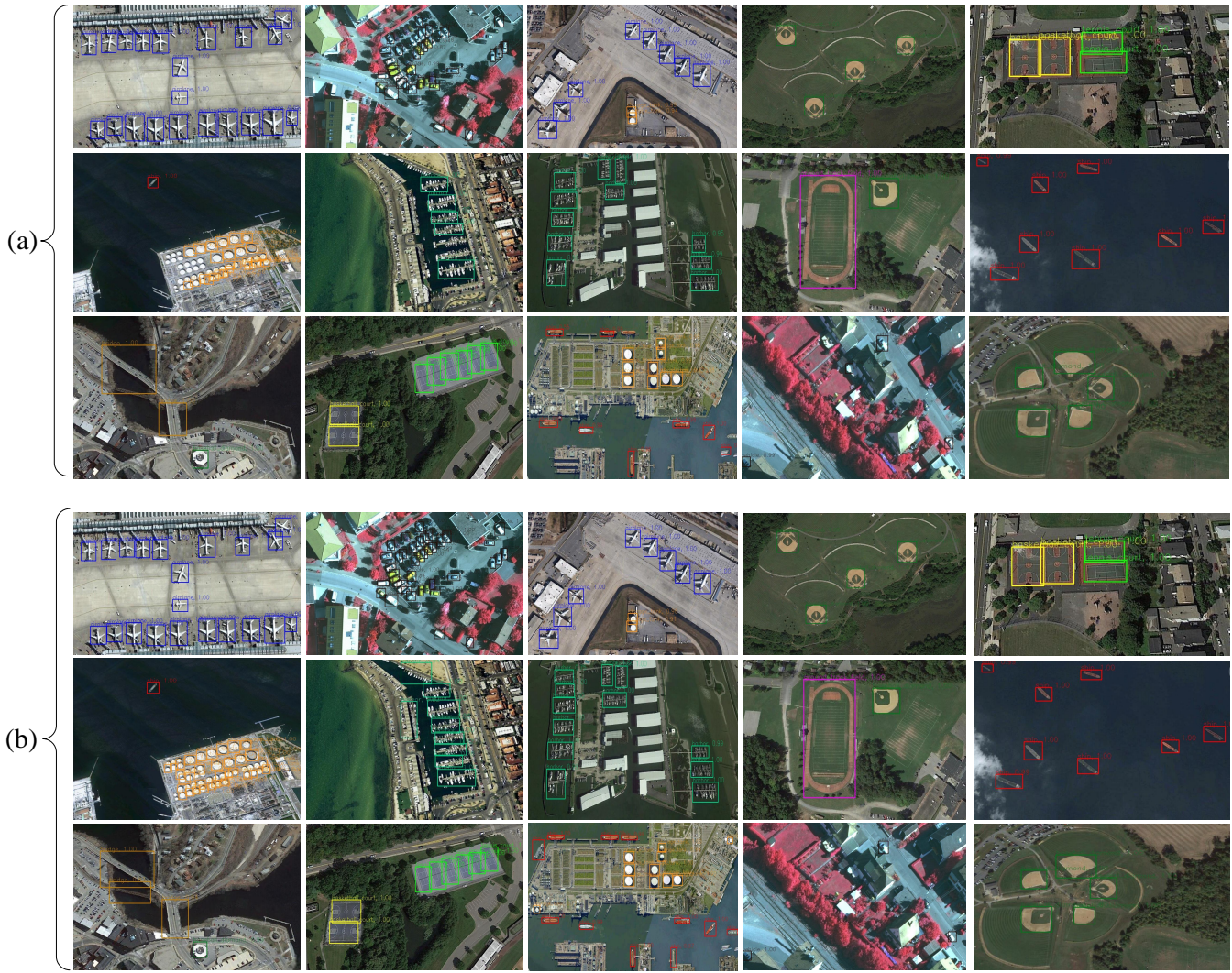}}
  \vspace{0cm}
\end{minipage}
\caption{Comparison of the selected detection results (the threshold is 0.4) for the Faster R-CNN and DAPNet methods. Only high-scoring detections are shown. (a) Detection results of the Faster R-CNN method. (b) Detection results of the DAPNet.}
\label{s5}
\end{figure*}

\begin{figure*}[htbp]
\centering
\begin{minipage}[b]{0\linewidth}
\centering
\centerline{\hspace{0cm}\epsfig{figure= 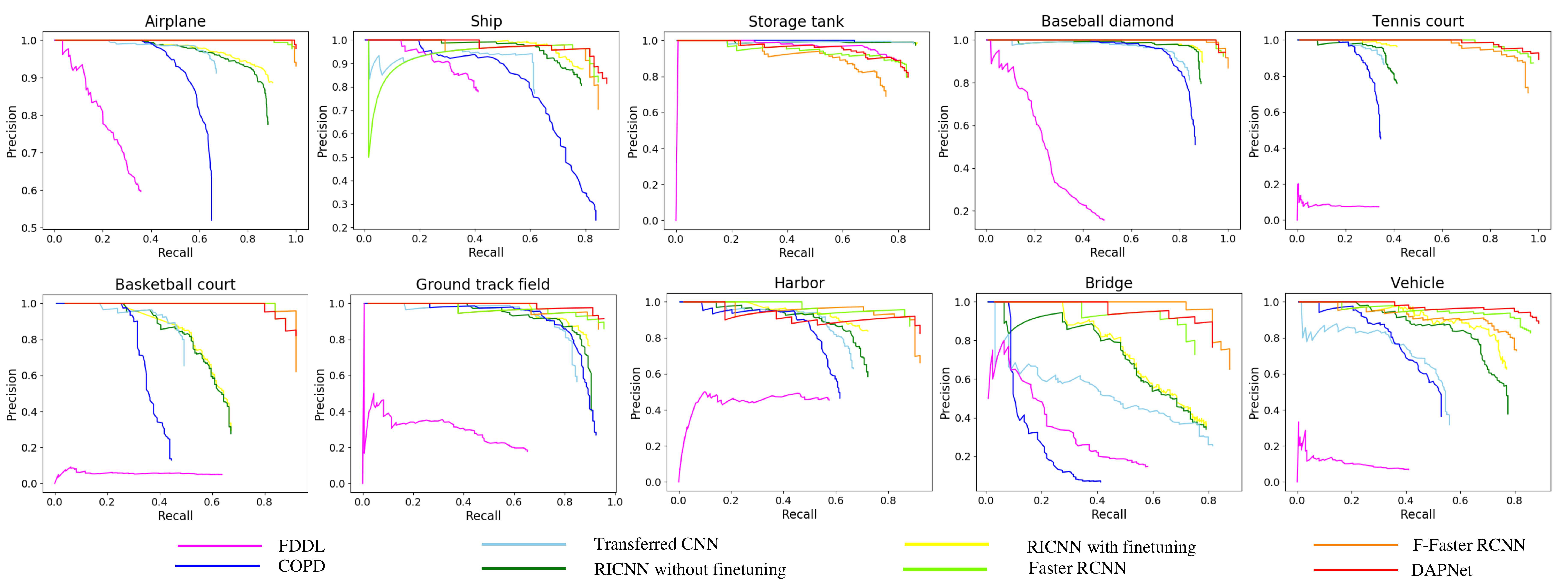,width=18.0cm}}
  \vspace{0cm}
\end{minipage}
\caption{PRCs of the proposed DAPNet and other state-of-the-art approaches for airplane, ship, storage tank, baseball diamond, temmis court, basketball court, ground track field, harbor, and vehicle categories, respectively.}
\label{s6}
\end{figure*}

\subsubsection{Region Proposal with F-RPN}
In experiments, the architecture of F-RPN is similar to RPN. However, the F-RPN produces plenty of proposal regions and the detail category confidences for each proposal region, and thus the specific class for each region proposal can be obtained by F-RPN. According to the category priors predicted by CPN and the positive factor, the adaptive candidate boxes are produced. For example, if the positive factor $pos=100$ and a category number is 3, we retain the 3$\times pos=3\times 100=300$ candidate boxes for this category. Besides, in order to insure the searching space for A-RCNN, the base number of candidate boxes for each category is 100. Compared with original Faster R-CNN, the number of candidate boxes for each image is fixed 2000. However, the DAPNet uses the F-RPN and CPN to adjust the number of candidate boxes, in line with the dense or spare characteristic of objects. Finally, the DAPNet naturally achieves the goal of reducing computation and improving accuracy.

The number of candidate boxes is the key point to limit the speed of two-stage detectors, and directly affect the recall rate of detection methods. Therefore, several groups of contrastive experiments are done by changing the positive factor. Fig. \ref{s8} explores the tradeoff between the proposed DAPNet method performance (measured with the Mean AP) and the quantity of the object proposals (measured with the positive factor) on our test data set. During the train process, the other hyper-parameters maintain fixed except for the positive factor. Hence, we derive the following from the Fig. \ref{s8}: 1) The mean AP improves rapidly and then tend to be stable with the increase of positive factor. 2) It is obvious that there is an eminent gap between the mean AP of object region proposals and the final detections. This gap proves the necessity of A-RCNN.

\renewcommand\arraystretch{1.5}
\begin{table*}[htbp]
\captionsetup{justification=centering}
\caption{\\PERFORMANCE COMPARISONS OF EIGHT DIFFERENT METHODS IN TERMS OF AP VALUES.\\
THE BOLD NUMBERS DENOTE THE HIGHEST VALUES IN EACH ROW}
\label{tab:threesome}
\centering
\begin{tabularx}{18cm}{lccXXXXXcccc}
    \hline
    \hline
    Method & airplane & ship & storage tank & baseball diamond & tennis court & basketball court & ground track field & harbor & bridge & vehicle & \textbf{mAP} \\
    \hline
    FDDL  & 0.2934 & 0.3768 & 0.7714 & 0.2584 & 0.0269 & 0.0361 & 0.2004 & 0.2541 & 0.2163 & 0.0436 & 0.2477  \\
    COPD  & 0.6301 & 0.7027 & 0.6580 & 0.8208 & 0.3351 & 0.3407 & 0.8527 & 0.5606 & 0.1564 & 0.4412 & 0.5499  \\
    Transferred CNN  & 0.6617 & 0.5709 & 0.8493 & 0.8174 & 0.3506 & 0.4611 & 0.7954 & 0.6224 & 0.4265 & 0.4305 & 0.5986  \\
    RICNN without finetuning  & 0.8617 & 0.7581 & 0.8502 & 0.8758 & 0.3927 & 0.5797 & 0.8579 & 0.6649 & 0.5834 & 0.6811 & 0.7106  \\
    RICNN with finetuning  & 0.8853 & 0.7789 & \textbf{0.8573} & 0.8857 & 0.4072 & 0.5780 & 0.8694 & 0.6804 & 0.6182 & 0.7151 & 0.7276  \\
    Faster RCNN & 0.9091 & 0.8021 & 0.7790 & 0.9091 & 0.9027 & 0.8182 & 0.8859 & 0.8031 & 0.7038 & 0.7882 & 0.8301  \\
    F-Faster RCNN & 0.9947 & 0.8053 & 0.6893 & 0.9925 & 0.8926 & \textbf{0.9053} & 0.8856 & \textbf{0.8818} & \textbf{0.8148} & 0.7570 & 0.8619\\
    DAPNet & \textbf{0.9990} & \textbf{0.8075} & 0.7888 & \textbf{0.9091} & \textbf{0.9870} & 0.8956 & \textbf{0.9026} & 0.8564 & 0.7935 & \textbf{0.8028} & \textbf{0.8742} \\
    \hline
    \hline
\end{tabularx}
\end{table*}
\normalsize

\subsection{Experimental Results and Comparisons}
Visual results of the experiments are shown in Fig. \ref{s5}, which are obtained by the proposed DAPNet framework and original Faster R-CNN on the test image of the NWPU VHR-10 dataset. The text on the left-upper of the rectangle denotes the object category and the predicted score for this rectangle box. Fig. \ref{s5}(a) is the detection results of the Faster R-CNN model, and Fig. \ref{s5}(b) is the detection results of the proposed DAPNet method. As listed in Fig. \ref{s5},
the Faster R-CNN method suffers from many missed detections, especially for dense objects, such as storage tank, and vehicle. In opposite, despite the large variations in the distribution and size of objects, the proposed approach has successfully detected most of the objects.

To comprehensive evaluate the effectiveness and superiority of the proposed DAPNet model, we compare it with the following methods: 1) The fisher discrimination dictionary learning (FDDL). 2) The collection of part detector (COPD). 3) A transferred CNN model from AlexNet \cite{Krizhevsky2012ImageNet}. 4) A newly trained rotation-invariant convolutional neural network (RICNN) with only the rotation-invariant layer and the softmax layer being trained (without fine-tuning the other layers). 5) A newly trained RICNN model (RICNN with fine-tuning) with all layers being trained. 6) A two-stage detection method Faster R-CNN.

In addition, to understand DAPNet better, we conduct ablation experiments to examine how each proposed component affects the final performance. We evaluate the performance of our method under two different settings: 1) F-Faster R-CNN: it only uses the F-RPN component based on the Faster R-CNN model, and ablates the CPN component. Comparison of the results of F-Faster R-CNN and Faster R-CNN, the effectiveness of F-RPN component can be shown. 2) DAPNet: it is our complete model, consisting of the CPN and the F-RPN components. The superiority of CPN is verified by observing the results of DAPNet and F-Faster R-CNN.

To make a fair comparison: 1) The same training data set and testing data set are used for the proposed DAPNet method and other comparison methods. 2) We uniformly employ the same not improved region proposal network parameter for Faster R-CNN, the presented F-Faster R-CNN, and the proposed DAPNet method to generate object proposals. 3) Our proposed DAPNet model, the F-Faster R-CNN model, the Faster R-CNN model, and the RICNN model are based on the VGG-16 model, and trained with the same weights that pretraind on $ImageNet$ $1000$-$class$ classification dataset.

\begin{figure}[htbp]
\centering
\begin{minipage}[b]{0\linewidth}
\centering
\centerline{\hspace{0cm}\epsfig{figure= 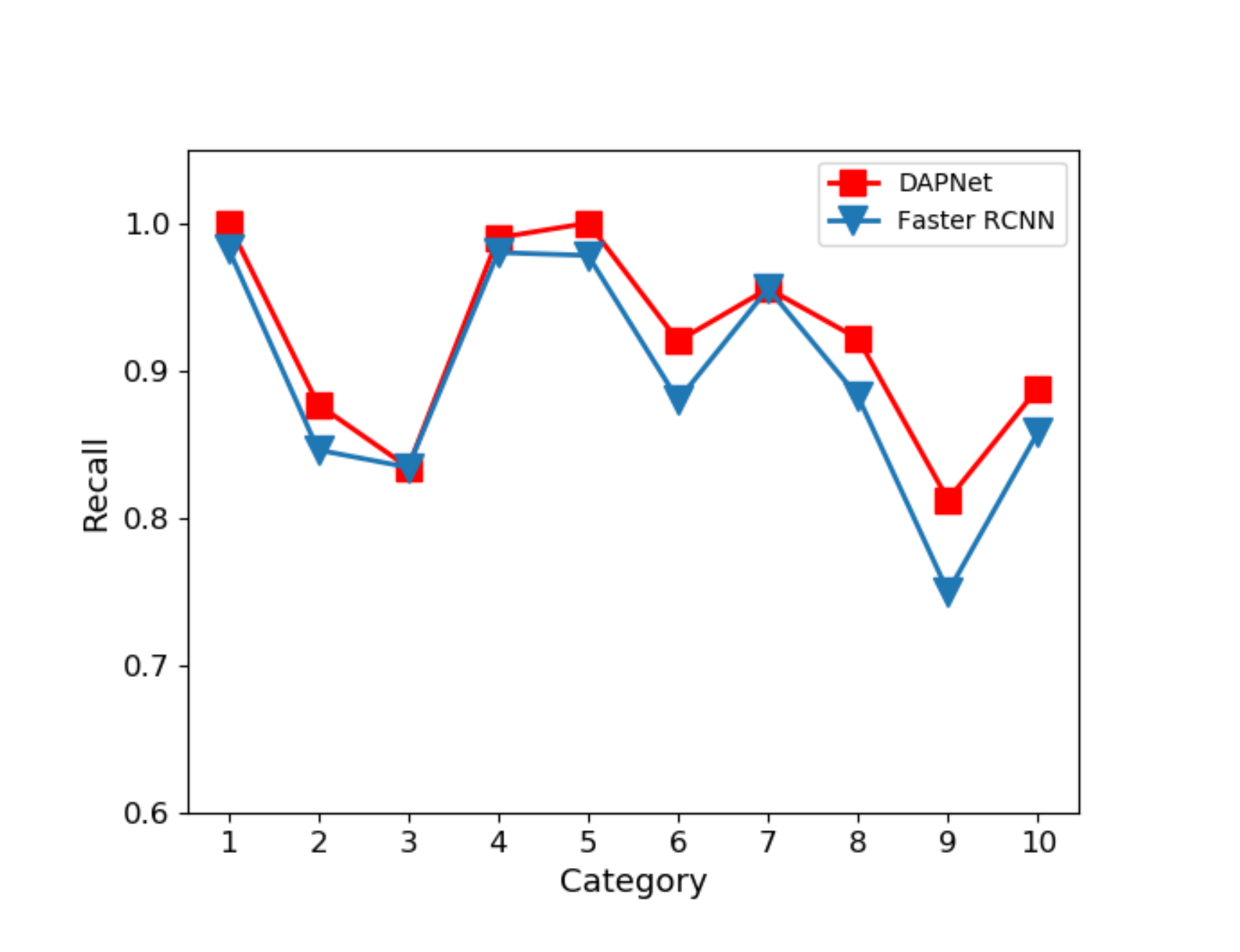, width=7.0cm}}
\end{minipage}
\caption{Quantitative evaluation results measured by recall rate for all 10 object categories (1-airplane, 2-ship, 3-storage tank, 4-baseball diamond, 5-tennis court, 6-basketball court, 7-ground track field, 8-harbor, 9-bridge, and 10-vehicle). The recall value obtained with Faster R-CNN and DAPNet.}
\label{s7}
\end{figure}

Fig. \ref{s7} reports the qualitative evaluation results of ten object categories for the proposed DAPNet and the Faster R-CNN algorithms. It can be seen that the proposed DAPNet algorithm demonstrates better detection performance on the ten classes, especially for the ship, basketball court, harbor, bridge and vehicle object categories. However, the proposed DAPNet indicates a less improvement on the object class of storage tank and ground track field. This can be easily explained: the DAPNet generates adaptive region proposals, while the Faster R-CNN produces 2000 region proposals for each image. Therefore, the number proposals for DAPNet is much less compared with Faster R-CNN for spare objects, and the Faster R-CNN reflects the same performance with DAPNet. In contrast, this phenomenon proves that our DAPNet can obtain equivalent performance with less region proposals.

Table \ref{tab:times} summarizes the computation cost of eight different methods. It shows that with the fewer computation cost, our DAPNet model improves significantly the overall detection precision. This adequately shows the effectiveness of the proposed DAPNet model learning method.

\renewcommand\arraystretch{1.1}
\begin{table}[htbp]
\captionsetup{justification=centering}
\caption{\\COMPUTATION TIME COMPARISONS OF EIGHT DIFFERENT METHODS}
\label{tab:times}
\centering
\begin{tabularx}{9cm}{l|c}
    \hline
    \hline
    Methods & Average running time per image (second) \\
    \hline
    FDDL & 7.54 \\
    COPD & 1.19 \\
    Transferred CNN & 5.37 \\
    RICNN without fine-tuning & 8.83 \\
    RICNN with fine-tuning & 8.83 \\
    Faster R-CNN & 0.289 \\
    F-Faster R-CNN & 0.382 \\
    DAPNet & 0.408 \\
    \hline
    \hline
\end{tabularx}
\end{table}
\normalsize

Table \ref{tab:threesome} and Fig. \ref{s6} show the quantitative comparison results of eight different methods, measured by AP values, and PRCs, respectively. As can be seen from them: 1) The proposed DAPNet method outperforms all other comparison approachs for all ten object classes in terms of mean AP.
Specially, our DAPNet methods obtained $8.99\%$, $8.43\%$, $7.74\%$, $5.33\%$, $8.97\%$ performance gains in terms of mean AP over the airplane, tennis
court, basketball court, harbor, and bridge, compared with the Faster R-CNN model, respectively.
This demonstrates the high superiority of the proposed method compared with the existing state-of-the-art object detection methods in remote sensing images. 2) By adding the category prior network, the performance measured in mean AP is further boosted by $1.23\%$, especially for the storage tank and the vehicle. However, our method has achieved the best performance, the detection accuracy for the object category of storage tank is lower than RICNN and transferred CNN. This is mainly due to the small size of storage tank category, the region proposal network was according to the high-level convolution features, while the selective search in RICNN produces region proposals on the basis of original images, and thus RICNN can obtained higher AP in small size and characteristic features compared with feature based region proposal networks. In summary, the experiments show the superiority of the proposed DAPNet model.

\section{Conclusion and future work}

In this paper, an novel and effective DAPNet framework is proposed to adapt the dense and sparse objects in optical remote sensing images and further improve the detection quality. The framework uses the CPN to predict category information for each class to as the prior information of F-RPN, combining the output region proposala of F-RPN, achieving the adaptive proposal network for each image.
The experiments demonstrate that our three contributions lead DAPNet to the state-of-the-art performance on a publicly available ten-class VHR object detection data set, especially for small objects.
However, the F-RPN based on the high-level convolution features that limits the scale of objects, in particular for the small objects with distinct feature information, such as the storage tank. And thus the traditional selective search shows the better performance compared with F-RPN in DAPNet.
Hence, in our future work, we intend to future improve the accuracy of small objects by learning a scale adaptation network.

\section{Acknowledgment}
The authors would like to thank the anonymous reviewers for their helpful comments.
Meanwhile, the authors would also like to thank Prof. Han team open the NWPU VHR-10 dataset.
\ifCLASSOPTIONcaptionsoff
  \newpage
\fi


\bibliographystyle{IEEEtran}
\bibliography{IEEEabrv,ref}

\end{document}